\documentclass[sigconf]{acmart}
\AtBeginDocument{%
  }

\copyrightyear{2026}
\acmYear{2026}
\setcopyright{cc}
\setcctype{by}
\acmConference[CIKM '26]{Proceedings of the 35th ACM International Conference on Information and Knowledge Management}{November 07--11, 2026}{Rome, Italy}
\acmBooktitle{Proceedings of the 35th ACM International Conference on Information and Knowledge Management (CIKM '26), November 07--11, 2026, Rome, Italy}
\acmDOI{10.1145/3799682.3840187}
\acmISBN{979-8-4007-2539-5/2026/11}

\usepackage{graphicx} 
\usepackage{subcaption}
\usepackage{url}
\usepackage{hyperref}
\usepackage{tcolorbox}
\usepackage{xcolor}
\usepackage{comment}
\usepackage{multirow}
\usepackage{adjustbox}
\usepackage{booktabs}
\usepackage{ulem}
\usepackage{subcaption}
\usepackage{svg}
\usepackage{mathrsfs}
\usepackage{mdframed}
\usepackage{listings}
\usepackage{algorithm}
\usepackage{amsmath}

\newcounter{todocounter}

\begin{document}

\title{WiCleanData: Guaranteeing the Type Consistency of Wikidata by Taxonomy Refinement and Constraint Enforcement}


\author{Yiwen Peng}
\orcid{0009-0007-7902-4097}
\affiliation{%
  \department{Télécom Paris}
  \institution{Institut Polytechnique de Paris}
  \city{Palaiseau}
  \country{France}
}
\email{yiwen.peng@telecom-paris.fr}

\author{Marc Jeanmougin}
\orcid{0000-0002-0176-7129}
\affiliation{%
  \department{Télécom Paris}
  \institution{Institut Polytechnique de Paris}
  \city{Palaiseau}
  \country{France}
}
\email{marc.jeanmougin@telecom-paris.fr}

\author{Thomas Bonald}
\orcid{0000-0003-0468-0384}
\affiliation{%
  \department{Télécom Paris}
  \institution{Institut Polytechnique de Paris}
  \city{Palaiseau}
  \country{France}
}
\email{thomas.bonald@telecom-paris.fr}





\renewcommand{\shortauthors}{Yiwen Peng, Marc Jeanmougin, and Thomas Bonald}

\begin{abstract}
Because of its collaborative nature, Wikidata suffers from errors, inconsistencies, and excessive complexity,
 such as redundant classes, ambiguity between instances and classes,  wrong taxonomic paths, and type constraint violations. The manual curation of these issues is infeasible at scale.
 To address these challenges, we introduce {\it WiCleanData}, a refined version of Wikidata with a consistent taxonomy and free from   type constraint violations.
 Specifically, we have designed an automated pipeline that first cleans the taxonomy with language model assistance, then simplifies type constraints by hierarchical aggregation, and finally filters facts accordingly.
 The resulting knowledge graph, free from any type violation, 
 is made publicly available via a Web interface, enabling easy exploration and downstream applications.
 
\end{abstract}





\maketitle


\section{Introduction}

\begin{figure*}
    \centering

    \begin{subfigure}{0.44\textwidth}
        \centering
        \includegraphics[width=\linewidth]{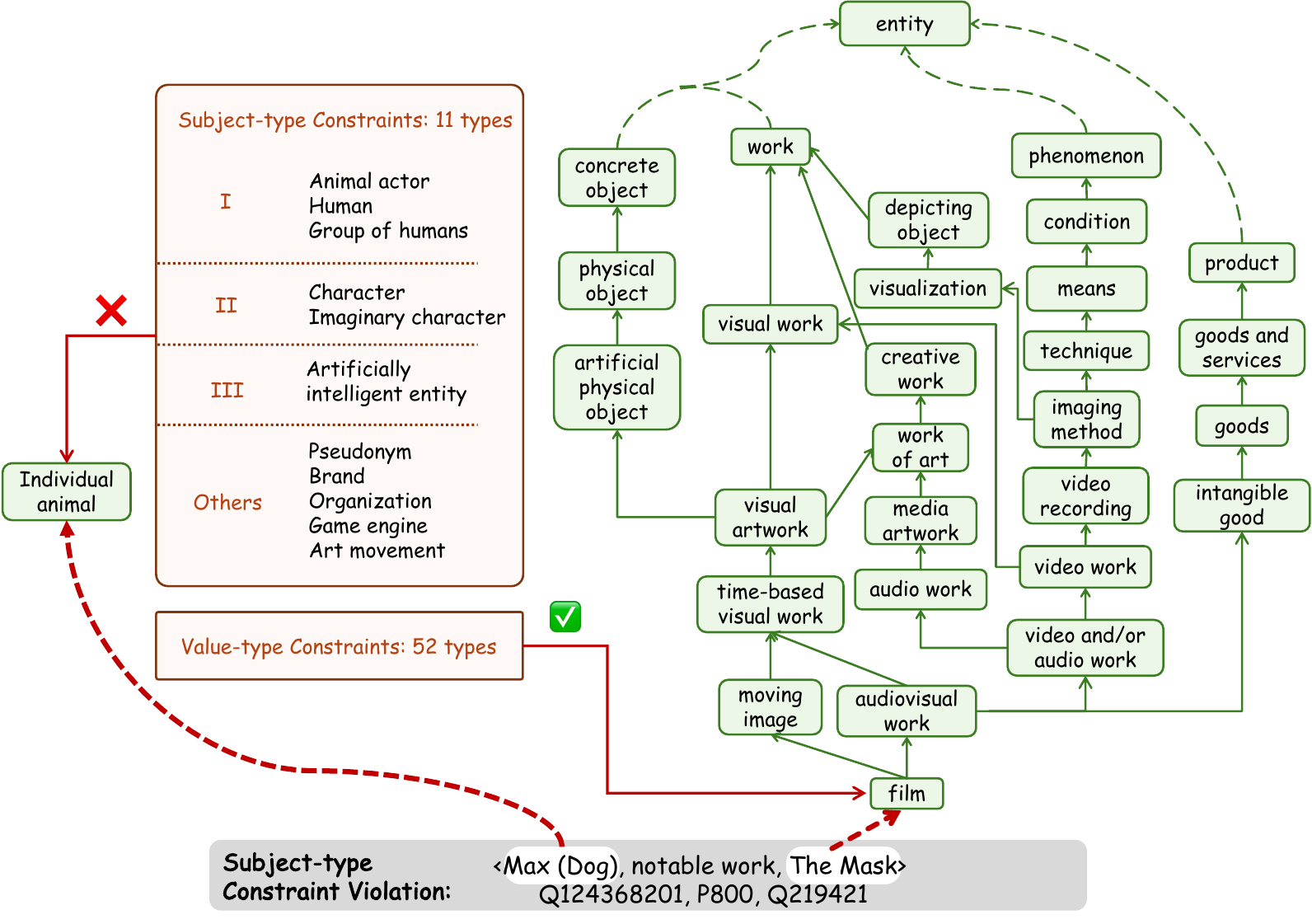}
        \caption{An excerpt of Wikidata}
        \label{fig:a}
    \end{subfigure}
    \hspace{0.03\textwidth}
    \begin{subfigure}{0.43\textwidth}
        \centering
        \includegraphics[width=\linewidth]{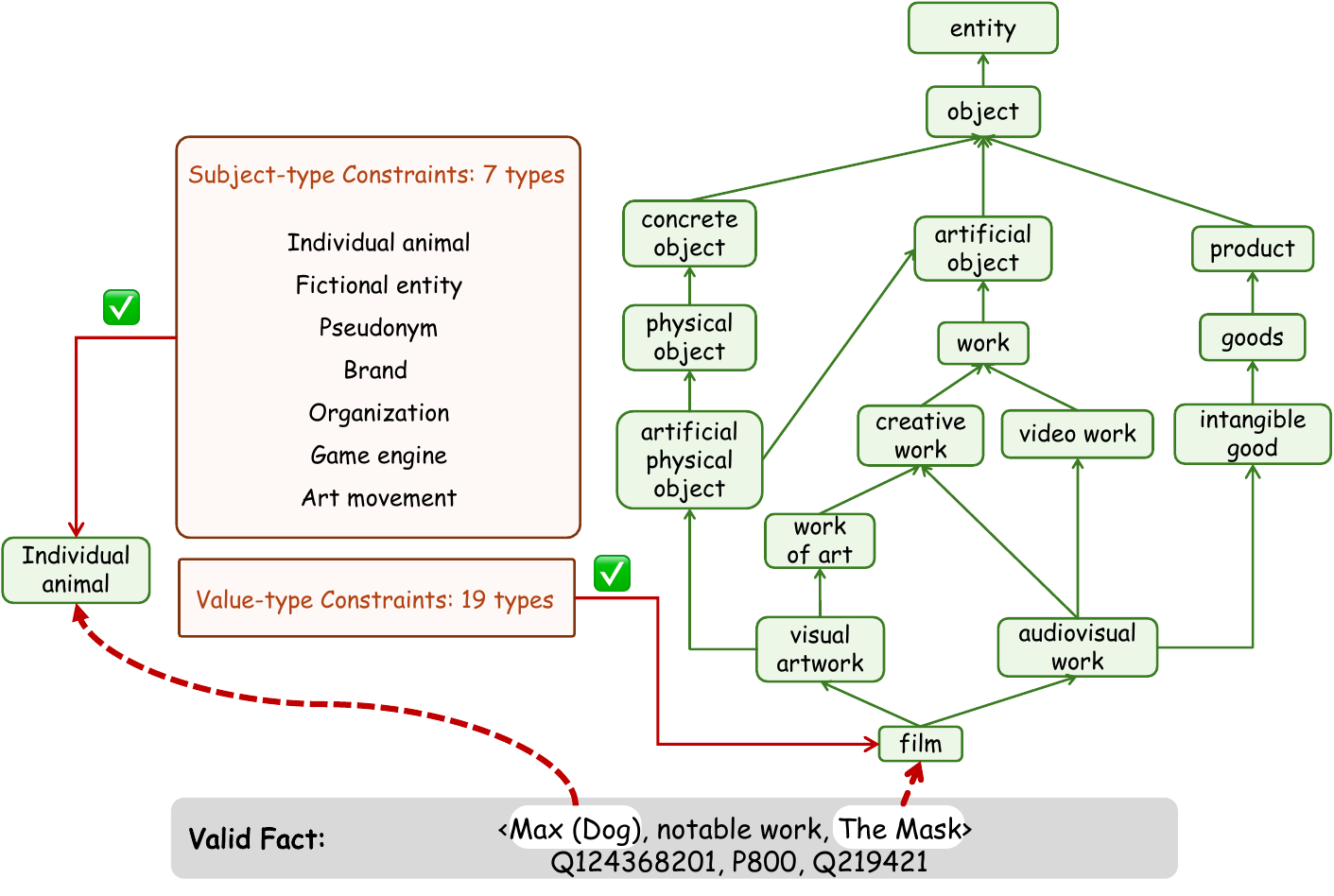}
        \caption{WiCleanData}
        \label{fig:b}
    \end{subfigure}
    \caption{Repairing type constraint violations of Wikidata by taxonomy cleaning.}
    \label{fig:all}
\end{figure*}

Wikidata is the largest and most widely used general-purpose Knowledge Graph (KG). Maintained by a large community of contributors, it serves as a core resource for a wide range of applications, including entity linking~\cite{raiman2018deeptype}, question answering~\cite{cao2022kqa}, and information retrieval~\cite{tran2022dense}.
However, due to its  collaborative nature, Wikidata suffers from errors, inconsistencies and excessive complexity.
For instance, as per Wikidata on May 9, 2026,
the class  \textit{film} is the 
 subclass of many redundant classes, including  \textit{visual work}, \textit{audiovisual work}, \textit{video and/or audio work}, \textit{video work}, 
  and has 
  \textit{condition} as one of its top-level classes (see Figure \ref{fig:a}). 
  The inherent complexity of the taxonomy  affects property type constraints, which are  difficult for contributors to verify and remain often incomplete.
For instance, the constrained subject types of the property \textit{notable work} include: (I) mixed class orders (e.g., the metaclass \textit{group of humans}), (III)  redundant types (e.g., \textit{imaginary character}, a subclass of {\it character}), and (II) orphan types with no instantiated facts of the corresponding property (e.g., \textit{artificially intelligent entity}).
  As a result, there are currently about 6M type-constraint violations in Wikidata, which is one of the major challenges for the community \cite{ferranti2025formalizing}. Many of these facts are actually valid, due to the incompleteness of the constraints. 
%
For instance, the fact \textit{Max, notable work, The Mask}, whose subject is an \textit{individual animal}, is incorrectly flagged as a type constraint violation (see Figure \ref{fig:a}). 

Addressing these issues is challenging due to the scale and the dynamic nature of Wikidata. Its truthy version exceeds 985GB and contains billions of facts  that are continuously updated by a large community of editors, tending to further increase the complexity of the 
taxonomy.
%
In this paper, we present WiCleanData, a refined version of Wikidata with a consistent taxonomy and free from type constraint violations. Specifically, we have designed an automated pipeline that first cleans the  taxonomy of Wikidata using Large Language Models (LLMs),  then simplifies and refines the type constraints through hierarchical aggregation  on the cleaned taxonomy. Finally, the facts are filtered according to the new taxonomy and type constraints. The quality of the resulting knowledge graph is assessed in terms of complexity, conciseness, robustness, coverage, and usefulness. 
WiCleanData is   publicly available via  a Web interface\footnote{\url{https://wicleandata.r2.enst.fr/}}  
for easy exploration and access.

The paper is structured as follows. Section~\ref{sec:related_work} reviews related work, Section~\ref{sec:clean_pipeline} describes the automated pipeline used to get  WiCleanData from Wikidata, Section~\ref{sec:evalation} presents the evaluation, and Section~\ref{sec:conclusion} concludes the paper.

\section{Related Work} \label{sec:related_work}
\paragraph{\textbf{Knowledge Graph Refinement}} KG refinement  aims at improving the  quality of KGs by detecting and correcting errors~\cite{paulheim2016knowledge}. 
Most existing works  focus on the detection of errors~\cite{jia2019triple,liang2017graph,li2017knowledge,melo2020automatic,topper2012dbpedia}, while some methods address both detection and repair~\cite{pellissier2019learning,wu2023task}. 
For instance, \cite{jiang2018knowledge} uses crowdsourcing to verify Linked Data quality issues, while~\cite{arioua2018user,bergman2015qoco} requires user interaction to clean violations. KGClean~\cite{ge2020kgclean} uses active learning to identify incorrect triples and repair them with candidates of maximum propagation-power scores. More recently, \cite{dong2025refining} feeds KGs directly into LLMs to detect erroneous triples and suggest repairs.
%
However, these methods are too resource-intensive to scale to very large KGs like Wikidata.
Instead, \cite{wang2016error} generates error hyperlinks in Wikipedia by the LinkRank algorithm and then trains an SVM classifier for links correction, while  CoCKG~\cite{melo2017approach} exploits disambiguation links and string matching to fix incorrect triples caused by entity confusion. Specifically for Wikidata,
~\cite{pellissier2019learning} mines correction rules from crowdsourced historical edits and uses these rules to fix constraint violations. 
However, the above methods mainly focus on ABox (instances and facts) and overlook issues in TBox (taxonomies and constraints themselves). 
\cite{wu2023task} studies taxonomy pruning in noisy KGs, but still ignores the type constraints.
In this paper, we address this limitation by automatically refining the taxonomy and enforcing proper type constraints.

\paragraph{\textbf{Efforts on Wikidata}}
Wikidata \cite{vrandevcic2014wikidata} is a collaborative KG actively maintained by a large community, but  still suffering from taxonomy complexity, redundancies, and inconsistencies \cite{shenoy2022study, brasileiro2016applying}.
To address these issues, Wikidata manages its data quality through a property constraint model\footnote{\url{https://www.wikidata.org/wiki/Help:Property_constraints_portal}}, which detects inconsistencies by enforcing or prohibiting specific data patterns~\cite{pellissier2019learning}. Recently, \cite{ferranti2025formalizing} formally defines Wikidata’s repair patterns for each property constraint based on its historical edits. For example, as for the TBox part, the repair primarily involves constraint deletion or class hierarchy additions, especially solving violations of \textit{Value-Type} and \textit{Type} constraints. 
%
%
Since 2023, Wikidata has launched the WikiProject Ontology Cleaning Task Force
\footnote{\url{https://www.wikidata.org/wiki/Wikidata:WikiProject_Ontology/Cleaning_Task_Force}}, 
which 
documents problem classes, and coordinates contributors to fix high-impact ontology issues. 
However, Wikidata relies on contributor collaborations, making ontology cleanup a lengthy and incremental process requiring individual commitment and consensus. 
Other resources have also proposed cleaning Wikidata. For example,
YAGO 4.5~\cite{suchanek2024yago} attempts to clean Wikidata by 
manually mapping the upper taxonomy to Schema.org
and defining the disjointness classes.
DBpedia~\cite{auer2007dbpedia} is derived from Wikipedia (and more recently, Wikidata) with a manually curated ontology. Its  automatic extraction process is also prone to inconsistencies~\cite{farber2017linked,abian2017wikidata}.
More recently, WiKC~\cite{peng2024refining} proposes a cleaned version of Wikidata taxonomy but ignores constraints and fact-level cleaning.
Despite these efforts, Wikidata refinement remains largely manual and time-consuming.
In this paper, we introduce a cleaned version of Wikidata by automatically cleaning its taxonomy and constraints, and filtering facts accordingly.

\section{Automated Pipeline} \label{sec:clean_pipeline}

We use the truthy version of Wikidata, which contains the best non-deprecated rank for each property, as of  May 9, 2026. We exclude 
 the class {\it scholarly article}  that is   controversial~\cite{suchanek2024yago} and  contributes to more than one billion facts.
 Specifically, we remove all instances, facts, and transitive subclasses of {\it scholarly article}. 
We also remove all entities without English labels (the vast majority are  of type {\it star}, {\it galaxy} and {\it Wikimedia category}), as well as all 
 facts associated with an external identifier\footnote{\url{https://www.wikidata.org/wiki/Wikidata:External_identifiers}}.
We get around 50M entities and 450M facts. 

\subsection{Taxonomy Refinement} \label{sec:taxonomy}

\label{sec:hierarchyExtraction}

There is no clear specification for the taxonomy of Wikidata.
In principle, it is defined by the  {\it subclassOf} (P279) property. However, this property is often confused with the {\it instanceOf} (P31) property by Wikidata contributors. 
To extract the taxonomy, we treat an entity as an actual class if the entity either has   at least one subclass or at least one instance with both a label and description; we then build the corresponding taxonomy using the {\it subclassOf} property. 
Next, we filter out classes that have no cumulative instances (i.e., classes where neither the class itself nor any of its transitive subclasses have instances).
To reduce complexity, we also remove all  metaclasses,
i.e.,  instances of {\it metaclass} or any of its transitive subclasses, except for the important classes {\it profession}, {\it occupation}, {\it taxon}, and {\it protein}, which we link to their  closest parent classes in the taxonomy. 
Finally, we  exclude all instances of \textit{BFO class}, which correspond to abstract concepts imported from the Basic Formal Ontology, except for \textit{role} and \textit{occurrence}, which we add as  subclasses of the root  \textit{entity}. By performing a depth-first search (DFS) from the root entity,  we break any cycle, bypass any class without description,  and remove any node that is not a transitive subclass of  \textit{entity}. 

\paragraph{\textbf{(1) Link Checking.}} 
To refine the taxonomy, we first question the relevance of each link using LLMs, inspired by \cite{peng2024refining}.
Specifically, if $a$ is a subclass of $b$ in  the taxonomy, we ask the LLM to confirm whether $a$ is a subclass of $b$, given the labels and descriptions of $a$ and $b$; we also ask the LLM whether $b$ is a subclass of $a$, in a symmetric way. Using both answers, say the boolean variables $subclass(a,b)$ and $subclass(b,a)$,
we  deduce the following semantic relation for the  link from $a$ to $b$:
\begin{align*}
subclass(a,b) \wedge \neg subclass(b,a) &\Rightarrow confirmed(a,b)\\
subclass(a,b) \wedge subclass(b,a) &\Rightarrow equivalent(a,b) \\
\neg subclass(a,b) \wedge subclass(b,a) &\Rightarrow reversed(a,b) \\
\neg subclass(a,b) \wedge \neg subclass(b,a) &\Rightarrow irrelevant(a,b) 
\end{align*}
where $confirmed(a,b)$ means that the link is confirmed by the LLM, $equivalent(a,b)$ means classes $a$ and $b$ are equivalent and should be merged, $reversed(a,b)$ means that the link should be reversed, and $irrelevant(a,b)$ means that the link between $a$ and $b$ should be cut. 
A confidence score is attached to each answer, based on the probability of the token associated with the expected answer (True or False); the predicted semantic  relation is 
taken into account
if the answers to both questions are valid (i.e., either True or False) and the minimum of the confidence scores of both answers exceeds some threshold $\theta$.
To mitigate bias~\cite{chen2024more}, the final decision is made from majority voting over predictions of several LLMs.

\paragraph{\textbf{(2) Graph Operations.}} 
Given the link predictions of the  LLMs, we successively proceed with the following graph transformations. We first (1) cut {\it irrelevant} links whenever the resulting disconnected subgraph, if any, has at most $k$ nodes, where $k$ is a parameter of the algorithm.
We then (2) cut  {\it reversed} links  if this does not disconnect any node from the root; otherwise, we  merge the two classes. 
 Next, we (3) merge {\it equivalent} classes and remove the created transitive links, if any. 
Finally, we (4) iteratively exclude leaf classes that have no instance, and classes of depth at least 3 that do not have a
Wikipedia page; disconnected classes are then relinked to their closest parent  in the original taxonomy.

\paragraph{\textbf{Example}}
An example is shown in Figure~\ref{fig:all}. The  link from \textit{video work} (defined as "{work with a video component}") 
to \textit{video recording} (defined as "{technique of recording, copying and broadcasting of moving visual images}")
is predicted as \textit{irrelevant}. This  link is then cut, and so eliminates the inconsistent taxonomy path from \textit{film} to \textit{condition}. 
In addition, \textit{goods and services} is predicted to be equivalent to \textit{product} and is therefore merged into it. Other redundant classes, such as \textit{time-based visual work}, are excluded as they don't have a  Wikipedia page. 
The resulting taxonomy ${\mathcal{T}}$ is much simpler, more accurate, and  less redundant (see Section \ref{sec:evalation}).

\subsection{Type Constraints} \label{sec:constraint-cleaning}

Next, we address 
the confusion, redundancy, and orphan types that
exist in the type constraints
of Wikidata. For each property, we retrieve the declared lists of subject-side and value-side constraint
classes and simplify each side independently using our  taxonomy~$\mathcal{T}$. 
Specifically, we map each class to its representative in taxonomy $\mathcal{T}$, as induced by the successive merging steps, and  drop it otherwise (e.g., if it is a metaclass). We then eliminate redundancy: any class that already has one of its ancestors in the type constraint is dropped. Finally, we simplify the type constraint in three steps: (1) hierarchical clustering of the remaining classes, (2) selection of  Lowest Common Ancestors (LCAs) within each cluster, and (3) filtering of orphan types. 

\paragraph{\textbf{(1) Hierarchical Clustering.}}
We cluster the   classes of each  type constraint as follows. We use the  distance between any two classes in the taxonomy ${\mathcal{T}}$ as a measure of 
 dissimilarity.
We then apply agglomerative hierarchical clustering with average linkage on the resulting
pairwise dissimilarity matrix and cut the dendrogram at the level corresponding to the largest gap
between consecutive merge heights. 
This cut produces a set of clusters  of classes, where each cluster corresponds to    classes that are close from one another in the taxonomy $\mathcal{T}$.

\paragraph{\textbf{(2) Lowest Common Ancestors.}}
For every cluster $\mathcal{C}$, let $\mathcal{L}$ be the 
set of all Lowest Common Ancestors (LCAs) of non-empty subsets of $\mathcal{C}$ in the taxonomy $\mathcal{T}$. We seek to cover $\mathcal{C}$ by some minimal subset of elements of $\mathcal{L}$ that are both close to $\mathcal{C}$ and do not introduce too many additional instances.
For each  $\ell \in \mathcal{L}$, let
$\mathrm{cov}(\ell)\subset \mathcal{C}$ be the subset of $\mathcal{C}$ covered by $\ell$.
For proximity, we keep only those $\ell \in \mathcal{L}$ whose 
the average distance  to their covered set $\mathrm{cov}(\ell)$ in the taxonomy $\mathcal{T}$ is less than some parameter $d$.
For representativeness, inspired by~\cite{10.5555/1625855.1625914}, we keep only $\ell \in \mathcal{L}$ whose proportion of additional cumulative instances (i.e., cumulative instances of $\ell$ that are not cumulative instances of $\mathrm{cov}(\ell)$) is less than some  parameter $\tau$.
Finally, we select a set $\mathcal{S}$ of minimum size    that covers $\mathcal{C}$, i.e., $\cup_{\ell\in \mathcal{S}} \mathrm{cov}(\ell) = \mathcal{C}$, among the remaining elements of $\mathcal{L}$.

\paragraph{\textbf{(3) Orphan Types.}} Finally, we remove orphan types from the constraint, i.e., any type that is not instantiated for this property after importing all valid facts (see Section~\ref{sec:factsEnrich}).

\paragraph{\textbf{Example.}} 
Consider the subject-type constraint of  property \textit{notable work}  in Figure~\ref{fig:all}. The classes \textit{human} and \textit{animal actor} are clustered to their LCA, \textit{individual animal}, thereby resolving the subject-type constraint violation. Other classes, such as \textit{group of humans} (a metaclass) or \textit{artificially intelligent entity} (an orphan type), are removed. The class \textit{character} is also removed as it is already covered by its ancestor  \textit{imaginary character}, which was merged into the class \textit{fictional entity} in the step of taxonomy cleaning.
%

\subsection{Fact Cleaning} \label{sec:factsEnrich}

Finally, we import facts from Wikidata using the cleaned taxonomy and type constraints. We proceed in two steps.

\paragraph{\textbf{(1) Instance Typing.}} We  collect instances and their type facts   from  Wikidata, based on the {\it instanceOf} (P31) property.
 We discard 
all type facts whose objects are not  classes of the cleaned  taxonomy ${\mathcal T}$. If the class has been filtered out in step (4) of the graph operations, we 
retype its instance to its closest ancestor, unless it is the root 
{\it entity}.
%
To reduce redundancy, we also remove all type facts that can be  deduced from another type fact by transitivity.
%


\paragraph{\textbf{(2) Constraint Filtering.}}
We then import all facts that meet the subject or value-type constraints derived above. Specifically, a triple $(s, p, o)$ whose property $p$ has either a  subject or object type constraint is imported whenever at least one direct type of the subject $s$ is a transitive subclass of one of the admissible subject classes of $p$, and likewise for the object  $o$.
Those facts whose properties have no type constraints are also imported.

\paragraph{\textbf{Results.}}
The resulting knowledge graph,  WiCleanData, has around 50M entities and 420M facts, representing about 100\% and 93\%, respectively, 
of those extracted from Wikidata (excluding scholarly articles, in particular). 
In other words, we retain nearly all entities but exclude 7\% of the facts due to missing or incorrect types.
By design, all facts of WiCleanData satisfy the type constraints.

The source code\footnote{\url{https://github.com/peng-yiwen/wicleanData}}  is publicly available for reproducibility. 
The automated pipeline was applied using three LLMs of different groups (Mistral-Small-24B, Qwen3-32B, and Gemma-3-27B) and the following parameters: $\theta=0.5$, 
$k=3, d=2, \tau=0.02$. 
A Web interface\footnote{\url{https://wicleandata.r2.enst.fr/}} 
is  provided to facilitate exploration of the knowledge graph. It  includes a SPARQL query endpoint, tools to compare the taxonomy to that of Wikidata and  to explain how the type constraints are generated for each property.


\section{Evaluation} \label{sec:evalation}

We now assess the quality of WiCleanData from both intrinsic and extrinsic perspectives.

\paragraph{ \textbf{Intrinsic evaluation.}}
Following~\cite{suchanek2024yago,unterkalmsteiner2023compendium}, we evaluate WiCleanData in terms of complexity, conciseness, readability, coverage, and robustness. 
The results are given in Table~\ref{tab:quality}. 
As expected, WiCleanData has a simpler, more concise taxonomy and fewer constraint types than Wikidata.
It also has fewer classes per instance, as redundancy has been reduced. The larger number of facts per instance is mainly due to instances of {\it scholarly article} and facts with external identifiers, which we have filtered out. 
We note that some classes in WiCleanData still have no direct instances. These classes are typically in the  upper part of taxonomy, such as \textit{spatial boundary}, having 
subclasses  like \textit{coastline} with  
instances. 
For robustness, we use the metric in \cite{unterkalmsteiner2023compendium} that  reflects the ability of a taxonomy to distinguish between different instances. It is typically inversely related  to conciseness, as a large number of classes makes entities easier to discriminate.
We observe that WiCleanData nearly preserves the taxonomic robustness, while substantially reducing redundancy.

\begin{table}
    \centering
    \caption{Intrinsic Evaluation}\label{tab:quality}
    \begin{adjustbox}{max width=0.45\textwidth}
    \begin{tabular}{llrr}
    \toprule
         Criterion & Metric &  ~~~~Wikidata &  ~~~~WiCleanData \\
         \midrule
         Memory & Dump size & 985G & \textbf{19G} \\
         Complexity & 
         Classes & 4.2M & \textbf{35K}\\
       &  Top-level classes & 30 &  \textbf{18} \\
                    & Metaclasses & 38K & \textbf{0} \\
                    &  Depth of the taxonomy & 22 & \textbf{14} \\
                    & Avg. number of paths to root & 75 & \textbf{5} \\  
                    & Avg. subject type constraints & 2.5 & \textbf{1.8} \\
                    & Avg. value type constraints & 3.6 & \textbf{2.4} \\
         Conciseness & Taxonomic loops & 36 & \textbf{0} \\                      
         & Redundant taxonomic links & 511K & \textbf{0} \\  
                     & Classes without instances & 93\% & \textbf{3\%} \\        
         Readability~~~ & Entities without label  & 24\% & \textbf{0\%} \\
         Coverage    & Classes per instance & \textbf{46.8} & 13.2  \\    
                     & Facts per instance & \textbf{21.3} & 5.3 \\
         Robustness  & Taxonomy robustness  & \textbf{0.80} & 0.75 \\
         \bottomrule
    \end{tabular}
    \end{adjustbox}
\end{table}

\paragraph{\textbf{Extrinsic evaluation.}} 
We further evaluate WiCleanData on multiple downstream tasks to show its usefulness and coverage. Specifically, we use the KGrEaT~\cite{heist2023kgreat} framework, which integrates multiple datasets and evaluation metrics into a modular architecture to evaluate the extrinsic utility of KGs. The results are shown in Table~\ref{tab:extrinsic}.
WiCleanData achieves comparable or even better performance than Wikdiata on all tasks of different types, proving its usability and coverage despite removing a large number of classes and facts. 
Clustering performance is slightly lower in terms of ARI and NMI, likely due to the removal of metaclass and their instances (see Limitations (1)), which reduces minority-cluster detectability. 

\begin{table}[h]
\centering
\caption{Extrinsic Evaluation (PK = Precision for known entities, PA = Precision for  all entities).} \label{tab:extrinsic}
\begin{adjustbox}{max width=0.4\textwidth}
\begin{tabular}{llcc|cc}
\hline
Task Type & Metric & \multicolumn{2}{c|}{\textbf{Wikidata}} & \multicolumn{2}{c}{\textbf{WiCleanData}} \\
 & & PK & PA & PK & PA \\
\hline
Classification & Accuracy $\uparrow$ & \textbf{0.558} & 0.344 & 0.551 & \textbf{0.408} \\
Regression & RMSE $\downarrow$ & 0.648 & 1.243 & \textbf{0.645} & \textbf{1.219} \\
Clustering & ARI $\uparrow$ & \textbf{0.197} & 0.086 & 0.114 & \textbf{0.087} \\
           & NMI $\uparrow$ & \textbf{0.229} & \textbf{0.154} & 0.115 & 0.147 \\
 & Accuracy $\uparrow$ & 0.621 & 0.318 & \textbf{0.724} & \textbf{0.426} \\
Doc. Sim. & Spearman $\uparrow$ & 0.167 & 0.108 & \textbf{0.200} & \textbf{0.199} \\
 & Pearson $\uparrow$ & 0.258 & 0.126 & \textbf{0.273} & \textbf{0.270} \\
 & Harm. Mean $\uparrow$ & 0.197 & 0.116 & \textbf{0.230} & \textbf{0.228} \\
Recommend. & F1 $\uparrow$ & \textbf{0.013} & \textbf{0.001} & \textbf{0.013} & \textbf{0.001} \\
\hline
\end{tabular}
\end{adjustbox}
\end{table}

\paragraph{ \textbf{Limitations.}} (1) In this work, we remove all metaclasses to simplify the taxonomy, which leads to the loss of some specific classes (e.g., \textit{type of chemical entity}, \textit{class of anatomical entity}). In Wikidata, instances of these classes are not always classes (e.g., \textit{Celahin D}, which is a molecule). Note that the concept of class order in Wikidata is controversial and difficult for most contributors to understand and refine ~\cite{brasileiro2016applying,patel2024class}.
(2) While LLMs show high reliability in hierarchy discovery~\cite{sun2024large}, they remain error-prone.
Currently, our predictions rely  on a single prompt template, though more effective templates may exist. 
Instead of exploring various combinations of LLMs and prompt templates, an alternative approach is to use confidence scores and study the taxonomy sensitivity to some confidence threshold.
(3) Current LLM prompting considers only local subclass links and ignores long-path dependencies. For instance,  \textit{book}  is correctly considered as a subclass of \textit{communications media}, but is arguably not a transitive subclass of  \textit{state}.

\section{Conclusion} \label{sec:conclusion}

In this paper, we introduce WiCleanData, a refined version of Wikidata containing a clean taxonomy, simplified type constraints, and type-consistent facts. It is built via an automatic three-stage pipeline: LLM-based taxonomy refinement, 
type constraint cleaning via hierarchical aggregation, and constraint-based fact filtering for type consistency.
Intrinsic evaluation shows that WiCleanData is less complex, more concise, and remains taxonomically robust. Extrinsic evaluation uses WiCleanData as background knowledge for multiple downstream tasks, and shows that it improves performance over Wikidata, confirming its usefulness and coverage.
As for future work, it includes several directions. For example, improve metaclass management~\cite{brasileiro2016applying} and consider more advanced repair strategies such as weakening and completion\cite{li2024repairing}.
%



\section*{GenAI Usage Disclosure}
As described in Section \ref{sec:taxonomy}, we adopted generative AI as part of the taxonomy refinement pipeline to assess the semantic validity of taxonomy links. We also used ChatGPT for language editing to improve the clarity of the manuscript. GenAI tools (e.g., Claude Code) were additionally used to assist with writing a small portion of the code, which was carefully reviewed, tested, and verified by the authors.
Beyond these uses, GenAI tools were not used for experimental results, scientific claims, or performance analysis.
The authors take full responsibility for the content of this paper.

\balance
\bibliographystyle{ACM-Reference-Format}
\bibliography{mybiblio}


\end{document}